\documentclass{article}
\usepackage{ijcai18}
\usepackage{times}
\usepackage{xcolor}
\usepackage{soul}
\usepackage[utf8]{inputenc}
\usepackage[small]{caption}
\usepackage{times}
\usepackage{helvet}
\usepackage{courier}
\usepackage{url}
\usepackage{graphicx}
\usepackage{multirow}
\usepackage{algorithm, algpseudocode}
\usepackage{float}
\usepackage{amsmath}
\usepackage{relsize}
\usepackage{multicol,lipsum}
\newtheorem{definition}{Definition}
\newtheorem{example}{Example}

\def\gets{:=}

\newcommand{\idest}{{\it i.e.}}
\newcommand{\exemp}{{\it e.g.}}

\newcommand{\etal}{{\it et al.}}

\newcommand{\incp}[1]{\widetilde{\mathcal{#1}}}
\newcommand{\inc}[1]{\widetilde{#1}}

\usepackage[disable]{todonotes}
\newcommand{\rfp}[2][noinline]{\todo[#1,color=orange!80]{RAMON: #2}}
\newcommand{\frm}[2][noinline]{\todo[#1,color=red!80]{FRM: #2}}

\usepackage{flushend}

\title{Heuristic Approaches for Goal Recognition in Incomplete Domain Models}

\author{
Ramon Fraga Pereira
\vspace{1.5mm}
\and Felipe Meneguzzi \\
Pontifical Catholic University of Rio Grande do Sul (PUCRS), Brazil \\
\vspace{1.5mm}
Postgraduate Programme in Computer Science, School of Technology \\
\texttt{ramon.pereira@acad.pucrs.br} \\
\vspace{1.5mm}
\texttt{felipe.meneguzzi@pucrs.br} \\
}

\begin{document}

\maketitle

\begin{abstract}

Recent approaches to goal recognition have progressively relaxed the assumptions about the amount and correctness of domain knowledge and available observations, yielding accurate and efficient algorithms. 
These approaches, however, assume completeness and correctness of the domain theory against which their algorithms match observations: this is too strong for most real-world domains. 
In this paper, we develop goal recognition techniques that are capable of recognizing goals using \textit{incomplete} (and possibly incorrect) domain theories. 
We show the efficiency and accuracy of our approaches empirically against a large dataset of goal and plan recognition problems with incomplete domains. 

\end{abstract}

\section{Introduction}

Goal recognition is the problem of identifying the correct goal intended by an observed agent, given a sequence of observations as evidence of its behavior in an environment and a domain model describing how the observed agent generates such behavior. 
Approaches to solve this problem vary on the amount of domain knowledge used in the behavior, or plan generation, model employed by the observed agent~\cite{ActivityIntentPlanRecogition_Book2014}, as well as the level of observability and noise in the observations used as evidence~\cite{Sohrabi_IJCAI2016}. 
Recent work has progressively relaxed the assumptions about the accuracy and amount of information available in observations required to recognize goals~\cite{NASA_GoalRecognition_IJCAI2015,Sohrabi_IJCAI2016,PereiraMeneguzzi_ECAI2016,PereiraNirMeneguzzi_AAAI2017}.
Regardless of the type of domain model formalism describing the observed agent's behavior, all recent approaches assume that the planning domain models are correct and complete, restricting its application to realistic scenarios in which the domain modeler either has an incomplete or incorrect model of the behavior under observation. 

Specifically, real-world domains have two potential sources of uncertainty:
(1) ambiguity in how actions performed by agents are realized; and (2) ambiguity from how imperfect sensor data reports features of the environment. 
The former stems from an incomplete understanding of the action being modeled and requires a domain modeler to specify a number of alternative versions of the same action to cover the possibilities. 
For example, an action to turn on the gas burner in a cooker may or may not require the observed agent to press a spark button. 
The latter stems from imperfections in the way actions themselves may be interpreted from real-world noisy data, \exemp, if one uses machine learning algorithms to classify objects to be used as features (\exemp, logical facts) of the observations~\cite{Granada2017}, certain features may not be reliably recognizable, so it is useful to model a domain with such feature as optional.

In this paper, we develop heuristic approaches to goal recognition to cope with incomplete planning domain models~\cite{Nguyen_AIJ_2017} and provide five key contributions. 
First, we formalize goal recognition in incomplete domains (Section~\ref{sec:problemFormulation}) combining the standard formalization of \citeauthor{RamirezG_AAAI2010}~\shortcite{RamirezG_IJCAI2009,RamirezG_AAAI2010} for plan recognition and that of \citeauthor{Nguyen_AIJ_2017}~\shortcite{Nguyen_AIJ_2017}. 
Second, we adapt an algorithm from~\cite{Hoffmann2004_OrderedLandmarks} to extract \textit{possible} landmarks in incomplete domain models (Section~\ref{section:ExtractingLandmarks_IncompleteDomains}). 
Third, we develop a notion of \textit{overlooked} landmarks that we can extract online as we process (\textit{on the fly}) observations that we can use to match candidate goals to the multitude of models induced by incomplete domains.
Fourth, we develop two heuristics that account for the various types of landmark as evidence in the observations to efficiently recognize goals (Section~\ref{section:HeuristicGoalRecognition_IncompleteDomains}).
Finally, we build a new dataset for goal recognition in incomplete domains based on an existing one~\cite{PereiraNirMeneguzzi_AAAI2017,Pereira_Meneguzzi_PRDatasets_2017} 
by removing varying amounts of information from complete domain models and annotating them with possible preconditions and effects that account for uncertain and possibly wrong information (Section~\ref{section:ExperimentsAndEvaluation}). 
We evaluate the approaches on this dataset and show that they are fast and accurate for recognizing goals in complex incomplete domain models at most percentages of incompleteness.


\section{Problem Formulation}
\label{sec:problemFormulation}

\subsection{STRIPS Domain Models}

We assume that the agents being observed reason using planning domains described in the STRIPS~\cite{STRIPSFikes1971} domain model $\mathcal{D} = \langle \mathcal{R}, \mathcal{O} \rangle$, where: $\mathcal{R}$ is a set of predicates with typed variables. 
Grounded predicates represent logical values according to some interpretation as facts, which are divided into two types: positive and negated facts, as well as constants for truth ($\top$) and falsehood ($\bot$); $\mathcal{O}$ is a set of operators $op = \langle \mathit{pre}(op), \mathit{eff}(op) \rangle$, where $\mathit{eff}(op)$ can be divided into positive effects $\mathit{eff}^{+}(op)$ (the add list) and negative effects $\mathit{eff}^{-}(op)$ (the delete list). 
An operator $op$ with all variables bound is called an action and allows state change. 
An action $a$ instantiated from an operator $op$ is applicable to a state $S$ iff $S \models \mathit{pre}(a)$ and results in a new state $S' = (S / \mathit{eff}^{-}(a)) \cup \mathit{eff}^{+}(a)$.
A planning problem within $\mathcal{D}$ and a set of typed objects $Z$ is defined as $\mathcal{P} = \langle \mathcal{F}, \mathcal{A}, \mathcal{I}, G \rangle$, where: $\mathcal{F}$ is a set of facts (instantiated predicates from $\mathcal{R}$ and $Z$); $\mathcal{A}$ is a set of instantiated actions from $\mathcal{O}$ and $Z$; $\mathcal{I}$ is the initial state ($\mathcal{I} \subseteq \mathcal{F}$); and $G$ is a partially specified goal state, which represents a desired state to be achieved. 
A plan $\pi$ for 
$\mathcal{P}$ is a sequence of actions $\langle a_1, a_2, ..., a_n \rangle$ that modifies the initial state $\mathcal{I}$ into a state $S\models G$ in which the goal state $G$ holds by the successive execution of actions in a plan $\pi$.

\subsection{Incomplete STRIPS Domain Models}
\label{sec:incompleteSTRIPS}

The agent reasoning about the observations and trying to infer a goal has information described using the formalism of incomplete domain models from \citeauthor{Nguyen_AIJ_2017}~\shortcite{Nguyen_AIJ_2017}, defined as $\widetilde{\mathcal{D}} = \langle \mathcal{R}, \widetilde{\mathcal{O}} \rangle$. 
Here, $\widetilde{\mathcal{O}}$ contains the definition of incomplete operators comprised of a six-tuple $\inc{op} = \langle \mathit{pre}(\inc{op}), \widetilde{\mathit{pre}}(\inc{op}), \mathit{eff}^{+}(\inc{op}), \mathit{eff}^{-}(\inc{op}), \widetilde{\mathit{eff}}^{+}(\inc{op}), \widetilde{\mathit{eff}}^{-}(\inc{op}) \rangle$, where: $\mathit{pre}(\inc{op})$ and $\mathit{eff}(\inc{op})$ have the same semantics as in the STRIPS domain models; and \textit{possible} preconditions $\widetilde{\mathit{pre}}(\inc{op}) \subseteq \mathcal{R}$ that \textit{might} be required as preconditions, as well as $\widetilde{\mathit{eff}}^{+}(\inc{op}) \subseteq \mathcal{R}$ and $\widetilde{\mathit{eff}}^{-}(\inc{op}) \subseteq \mathcal{R}$ that \textit{might} be generated as \textit{possible} effects respectively as add or delete effects. 
An incomplete domain $\widetilde{\mathcal{D}}$ has a \textit{completion set} $\langle\langle \widetilde{\mathcal{D}} \rangle\rangle$ comprising all possible domain models derivable from the incomplete one. 
There are $2^K$ possible such models where $K = \sum_{\inc{op} \in \widetilde{\mathcal{O}}}(|\widetilde{\mathit{pre}}(\inc{op})| + |\widetilde{\mathit{eff}}^{+}(\inc{op})| + |\widetilde{\mathit{eff}}^{-}(\inc{op})|)$, and a single (unknown) ground-truth model $\mathcal{D}^{*}$ that actually drives the observed state. 
An incomplete planning problem derived from an incomplete domain $\incp{D}$ and a set of typed objects $Z$ is defined as $\incp{P} = \langle \mathcal{F}, \incp{\mathcal{A}}, \mathcal{I}, G \rangle$, where: $\mathcal{F}$ is the set of facts (instantiated predicates from $Z$), $\incp{\mathcal{A}}$ is the set of incomplete instantiated actions from $\inc{\mathcal{O}}$ with objects from $Z$, $\mathcal{I} \subseteq \mathcal{F}$ is the initial state, and $G \subseteq \mathcal{F}$ is the goal state.

Most approaches to planning in incomplete domains~\cite{WeberBryce_ICAPS_2011,PlanningIncomplete_NguyenK_2014,Nguyen_AIJ_2017} assume that plans succeed under the \textit{most optimistic} conditions, namely that: possible preconditions do not need to be satisfied in a state; possible add effects are always assumed to occur in the resulting state; and delete effects are ignored in the resulting state. 
Formally, an incomplete action $\inc{a}$ instantiated from an incomplete operator $\inc{op}$ is applicable to a state $S$ iff $S \models \mathit{pre}(\inc{a})$ and results in a new state $S'$ such that $S' \gets (S / \mathit{eff}^{-}(a)) \cup (\widetilde{\mathit{eff}}^{+}(\inc{a}) \cup \mathit{eff}^{+}(a))$. Thus, a valid plan $\pi$ that achieves a goal $G$ from $\mathcal{I}$ in an incomplete planning problem $\incp{P}$ is a sequence of actions that corresponds to an \textit{optimistic} sequence of states. 
Example~\ref{example:Abstract} from \citeauthor{WeberBryce_ICAPS_2011}~\shortcite{WeberBryce_ICAPS_2011} illustrates an abstract incomplete domain and a valid plan for it.

\begin{example}\label{example:Abstract}
Let $\incp{P}$ be incomplete planning problem, where:
	$\mathcal{F} = \lbrace p,q,r,g \rbrace$; 
	$\incp{\mathcal{A}} = \lbrace \inc{a},\inc{b},\inc{c} \rbrace$, where: 
	{\scriptsize
	\begin{itemize}
		\itemsep0em 
		\item $\mathit{pre}(\inc{a}) = \lbrace p,q \rbrace, \widetilde{\mathit{pre}}(\inc{a}) = \lbrace r \rbrace, \widetilde{\mathit{eff}}^{+}(\inc{a}) = \lbrace r \rbrace, \widetilde{\mathit{eff}}^{-}(\inc{a}) = \lbrace p \rbrace$
		\item $\mathit{pre}(\inc{b}) = \lbrace p \rbrace, \mathit{eff}^{+}(\inc{b}) = \lbrace r \rbrace, \mathit{eff}^{-}(\inc{b}) = \lbrace p \rbrace, \widetilde{\mathit{eff}}^{-}(\inc{b}) = \lbrace q \rbrace$
		\item $\mathit{pre}(\inc{c}) = \lbrace r \rbrace, \widetilde{\mathit{pre}}(\inc{c}) = \lbrace q \rbrace, \mathit{eff}^{+}(\inc{c}) = \lbrace g \rbrace$
		\item $\mathcal{I} = \lbrace p,q \rbrace$; and
	$G = \lbrace g \rbrace$.
	\end{itemize}
	}

The $[\inc{a},\inc{b},\inc{c}]$ sequence of actions is a valid plan to achieve goal state $\lbrace g \rbrace$ from the initial state $\lbrace p,q \rbrace$.
It corresponds to the \textit{optimistic} state sequence: $s_{0} =  \lbrace p, q \rbrace, s_{1} = \lbrace p,q,r \rbrace, s_{2} = \lbrace q,r \rbrace, s_{3} = \lbrace q, r, g \rbrace$. 
The number of completions for this example is $|\langle\langle \widetilde{\mathcal{D}} \rangle\rangle| = 2^{5}$ (2 possible preconditions and 3 possible effects, \idest, 1 possible add effect and 2 possible delete effects).
\end{example}

\subsection{Goal Recognition in Incomplete Domains}
\label{section:GoalRecognition_IncompleteDomains}

Goal recognition is the task of recognizing and anticipating agents' goals by observing their interactions in an environment. 
Whereas most planning-based goal recognition approaches assume complete domain model~\cite{RamirezG_IJCAI2009,RamirezG_AAAI2010,GoalRecognitionDesign_Keren2014,NASA_GoalRecognition_IJCAI2015,Sohrabi_IJCAI2016,PereiraMeneguzzi_ECAI2016,PereiraNirMeneguzzi_AAAI2017}, we assume that the observer has an incomplete domain model while the observed agent is planning and acting with a complete domain model. 
To account for such uncertainty, the model available to the observer contains possible preconditions and effects as defined in Section~\ref{sec:incompleteSTRIPS}. Like most planning approaches in incomplete domains~\cite{WeberBryce_ICAPS_2011,PlanningIncomplete_NguyenK_2014,Nguyen_AIJ_2017}, we reason about possible plans with incomplete actions (observations) by assuming that they succeed under the \textit{most optimistic} conditions. 
We formalize goal recognition over incomplete domain models in Definition~\ref{def:GoalRecognitionIncompleteDomains}. 

\begin{definition}[\textbf{Goal Recognition Problem}]
	\label{def:GoalRecognitionIncompleteDomains}
A goal recognition problem with an incomplete domain model is a quintuple $\incp{T} = \langle \incp{D}, Z, \mathcal{I}, \mathcal{G}, Obs \rangle$, 
where: 
\begin{itemize}
	\itemsep0em 
	\item $\incp{D} = \langle \mathcal{R}, \widetilde{\mathcal{O}} \rangle$ is an incomplete domain model (with possible preconditions and effects). 
	$Z$ is the set of typed objects in the environment, in which $\mathcal{F}$ is the set of instantiated predicates from $Z$, and $\incp{\mathcal{A}}$ is the set of incomplete instantiated actions from $\inc{\mathcal{O}}$ with objects from $Z$;
	\item $\mathcal{I} \in \mathcal{F}$ an initial state;
	\item $\mathcal{G}$ is the set of possible goals, which include a correct hidden goal $G^*$ (\idest, $G^*$ $\in$ $\mathcal{G}$); and
	\item $Obs = \langle o_1, o_2, ..., o_n\rangle$ is an observation sequence of executed actions, with each observation $o_i \in \incp{\mathcal{A}}$. 
	$Obs$ corresponds to the sequence of actions (\idest, a plan) to solve a problem in a complete domain in $\langle\langle \widetilde{\mathcal{D}} \rangle\rangle$.
\end{itemize}
\end{definition}

A solution for a goal recognition problem in incomplete domain models $\incp{T}$ is the correct hidden goal $G^* \in \mathcal{G}$ that the observation sequence $Obs$ of a plan execution achieves. 
As most goal recognition approaches, observations consist of the action signatures of the underlying plan\footnote{Our approaches are not limited to using just actions as observations and can also deal with logical facts as observations.}, more specifically, we observe incomplete actions with possible precondition and effects, in which some of the preconditions might be required and some effects might change the environment.
While a full (or complete) observation sequence contains all of the action signatures of the plan executed by the observed agent, an incomplete observation sequence contains only a sub-sequence of actions of a plan and thus misses some of the actions actually executed in the environment. 

\section{Landmark Extraction in Incomplete Domains}\label{section:ExtractingLandmarks_IncompleteDomains}

In planning, landmarks are facts (or actions) that must be achieved (or executed) at some point along all valid plans to achieve a goal from an initial state~\cite{Hoffmann2004_OrderedLandmarks}. 
Landmarks are often used to build heuristics for planning algorithms~\cite{LandmarksRichter_2008,RichterLPG_2010}. 
Whereas landmark-based heuristics extract landmarks from complete and correct domain models in the planning literature, we extend the landmark extraction algorithm of Hoffmann~\etal~in~\shortcite{Hoffmann2004_OrderedLandmarks} to extract \textit{definite} and \textit{possible} landmarks in incomplete STRIPS domain models.
This algorithm uses a Relaxed Planning Graph (RPG), which is a leveled graph that ignores the delete-list effects of all actions, thus containing no mutex relations~\cite{FFHoffmann_2001}. 
Once the RPG is built, this algorithm extracts a set of \textit{landmark candidates} by back-chaining from the RPG level in which all facts of the goal state $G$ are possible, and, for each fact $g$ in $G$, it checks which facts must be true until the first level of the RPG. 
For example, if fact $B$ is a landmark and all actions that achieve $B$ share $A$ as precondition, then $A$ is a landmark candidate. 
To confirm that a landmark candidate is indeed a necessary condition, and thus a landmark, the algorithm builds a new RPG removing actions that achieve the landmark candidate and checks the solvability over this modified problem. 
If the modified problem is unsolvable, then the landmark candidate is a necessary landmark. 
This means that the actions that achieve the landmark candidate are necessary to solve the original planning problem.  
Deciding the solvability of a relaxed planning problem using an RPG structure can be done in polynomial time~\cite{BlumFastPlanning_95}. 

We adapt the extraction algorithm from~\citeauthor{Hoffmann2004_OrderedLandmarks}~\shortcite{Hoffmann2004_OrderedLandmarks}\footnote{We chose this specific algorithm due to its simplicity and runtime efficiency.} to extract landmarks from incomplete domain models by building an Optimistic Relaxed Planning Graph (ORPG) instead of the original RPG. 
An ORPG is leveled graph that deals with incomplete domain models by assuming the most \textit{optimistic} conditions. 
Thus, besides ignoring the delete-effects of all actions, this graph also ignores possible preconditions and possible delete-effects, whereas we use all possible add effects. 
The ORPG allows us to extract \textit{definite} and \textit{possible} landmarks, formalized in Definitions~\ref{def:DefiniteLandmark} and~\ref{def:PossibleLandmark}.

\begin{definition}[\textbf{Definite Landmark}]\label{def:DefiniteLandmark}
A \textit{definite} landmark $L_{Definite}$ is a fact (landmark) extracted from a known add effect $\mathit{eff}^{+}(a)$ of an achiever\footnote{An achiever is an action at the level before a candidate landmark in the RPG that can be used to achieve this candidate landmark.} $a$ (action) in the ORPG.
\end{definition}

\begin{definition}[\textbf{Possible Landmark}]\label{def:PossibleLandmark}
A \textit{possible} landmark $L_{Possible}$ is a fact (landmark) extracted from a possible add effect $\widetilde{\mathit{eff}}^{+}(a)$ of an achiever $a$ (action) in the ORPG.
\end{definition}

\begin{figure}[h!]
  \centering
  \includegraphics[width=0.6\linewidth]{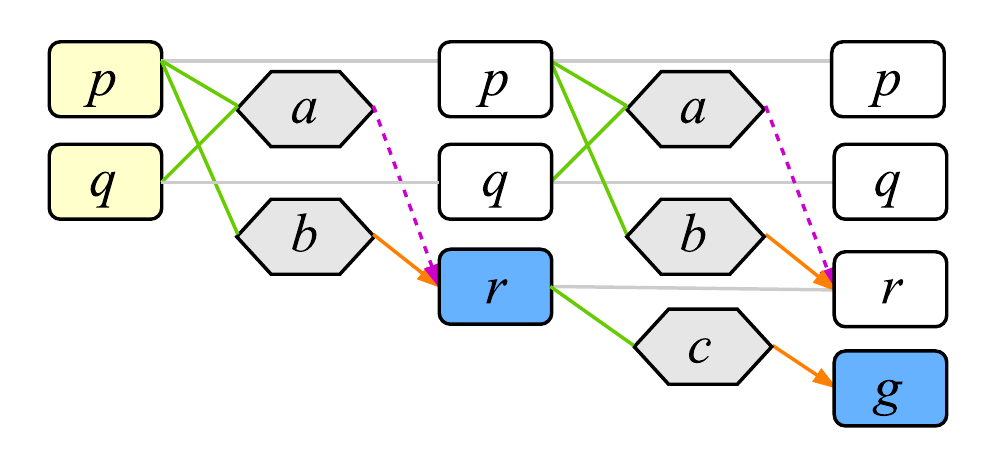}
  \caption{ORPG for Example~\ref{example:Abstract}. Green arrows represent preconditions, Orange arrows represent add effects, and Purple dashed arrows represent possible add effects. Light-Blue boxes represent the set of \textit{definite} landmarks and Light-Yellow boxes represent the set of \textit{possible} landmarks. Hexagons represent actions.}
  \label{fig:ORPG_Example}
\end{figure}

Figure~\ref{fig:ORPG_Example} shows the ORPG for Example~\ref{example:Abstract}.
The set of \textit{definite} landmarks is $\lbrace r,g \rbrace$ (Light-Blue in Figure~\ref{fig:ORPG_Example}), and the set of \textit{possible} landmarks is $\lbrace p,q \rbrace$ (Light-Yellow in Figure~\ref{fig:ORPG_Example}). 
The classical landmark extraction algorithm from~\citeauthor{Hoffmann2004_OrderedLandmarks} (without the most \textit{optimistic} conditions), returns $\lbrace p,r,g \rbrace$ as landmarks ignoring $q$ as a fact landmark because it does not assume the most \textit{optimistic} condition that possible add effects always occur. 
Therefore, the $a$ action was not considered as a possible achiever action. 
Although we use this modification to recognize goals in incomplete domain models, these landmarks can easily be used to build heuristics for planning in incomplete domains. 

\section{Heuristic Goal Recognition Approaches}\label{section:HeuristicGoalRecognition_IncompleteDomains}

\frm{Are overlooked landmarks from the set of definite landmarks? Or can possible preconditions/effects be overlooked landmarks?}
\rfp{It depends on two factors: if the fact is extracted from known add effects, then it is a definite landmark. However, if it is extracted from possible add effects, then it is an overlooked landmark. We just check the observed facts on-the-fly.}
Key to our goal recognition approaches are observing the evidence of achieved landmarks during observations to recognize which goal is more consistent with the observations. 
To do so, our approaches combine the concepts of \textit{definite} and \textit{possible} with that of \textit{overlooked} landmarks. An overlooked landmark is an actual landmark, \idest, a necessary fact for all valid plans towards a goal, that was not detected by approximate landmark extraction algorithms. 
Since we are dealing with incomplete domain models, and it is possible that they have few (or no) \textit{definite} and/or \textit{possible} landmarks, we extract \textit{overlooked} landmarks from the evidence in the observations as we process them in order to enhance the set of landmarks useable by our heuristic. 
To find such landmarks we build a new ORPG removing observed actions that achieve a potentially \textit{overlooked} fact landmark and check the solvability of this modified problem. 
If the modified problem is indeed unsolvable, then this fact is an \textit{overlooked} landmark. 

\subsection{Goal Completion Heuristic}

We now combine our notions of landmarks to develop a goal recognition heuristic for recognizing goals in incomplete domain models. 
Our heuristic estimates the correct goal in the set of candidate goals by calculating the ratio between achieved \textit{definite} ($\mathcal{AL}_{G}$), \textit{possible} ($\mathcal{\widetilde{AL}}_{G}$), and \textit{overlooked} ($\mathcal{ANL}_{G}$) landmarks and the amount of \textit{definite} ($\mathcal{L}_{G}$), \textit{possible} ($\mathcal{\widetilde{L}}_{G}$), and \textit{overlooked} ($\mathcal{NL}_{G}$) landmarks. 
The estimate computed using Equation~\ref{eq:Heuristic} represents the percentage of achieved landmarks for a candidate goal from observations. 

{
\begin{equation}
\label{eq:Heuristic}
h_{\widetilde{GC}}(G) = \left(\frac{\mathcal{AL}_{G} + \mathcal{\widetilde{AL}}_{G} + \mathcal{ANL}_{G}}{\mathcal{L}_{G} + \mathcal{\widetilde{L}}_{G} + \mathcal{NL}_{G}}\right)
\end{equation}
}
\subsection{Uniqueness Heuristic}

Most goal recognition problems contain multiple candidate goals that share common fact landmarks, generating ambiguity that jeopardizes the goal completion heuristic. 
Clearly, landmarks that are common to multiple candidate goals are less useful for recognizing a goal than landmarks that exist for only a single goal. 
As a consequence, computing how unique (and thus informative) each landmark is can help disambiguate similar goals for a set of candidate goals. 
Our second goal recognition heuristic is based on this intuition, which we develop through the concept of \textit{landmark uniqueness}, which is the inverse frequency of a landmark among the landmarks found in a set of candidate goals. 
Intuitively, a landmark $L$ that occurs only for a single goal within a set of candidate goals has the maximum uniqueness value of 1. 
Equation~\ref{eq:LandmarksUniqueness} formalizes the computation of the \textit{landmark uniqueness value} for a landmark $L$ and a set of landmarks for all candidate goals $K_{\mathcal{G}}$.

{
\begin{equation}
\label{eq:LandmarksUniqueness}
L_{\mathit{Uniq}}(L, K_{\mathcal{G}}) = \left(\frac{1}{\displaystyle\sum_{\mathcal{L} \in K_{\mathcal{G}}} |\{L |L \in \mathcal{L}\}|}\right)
\end{equation}
}

Using the concept of \textit{landmark uniqueness value}, we estimate which candidate goal is the intended one by summing the uniqueness values of the landmarks achieved in the observations. 
Unlike our previous heuristic, which estimates progress towards goal completion by analyzing just the set of achieved landmarks, the landmark-based uniqueness heuristic estimates the goal completion of a candidate goal $G$ by calculating the ratio between the sum of the uniqueness value of the achieved landmarks of $G$ and the sum of the uniqueness value of all landmarks of a goal $G$. 
Our new uniqueness heuristic also uses the concepts of \textit{definite}, \textit{possible}, and \textit{overlooked} landmarks. 
We store the set of \textit{definite} and \textit{possible} landmarks of a goal $G$ separately into $\mathcal{L}_{G}$ and $\mathcal{\widetilde{L}}_{G}$, and the set of \textit{overlooked} landmarks into $\mathcal{NL}_{G}$. 
Thus, the uniqueness heuristic effectively weighs the completion value of a goal by the informational value of a landmark so that unique landmarks have the highest weight. 
To estimate goal completion using the \textit{landmark uniqueness value}, we calculate the uniqueness value for every extracted (\textit{definite}, \textit{possible}, and \textit{overlooked}) landmark in the set of landmarks of the candidate goals using Equation~\ref{eq:LandmarksUniqueness}.
Since we use three types of landmarks and they are stored in three different sets, we compute the landmark uniqueness value separately for them, storing the landmark uniqueness value of \textit{definite} landmarks $\mathcal{L}_{G}$ into $\Upsilon_{\mathcal{L}}$, the landmark uniqueness value of \textit{possible} landmarks $\mathcal{\widetilde{L}}_{G}$ into $\Upsilon_{\mathcal{\widetilde{L}}}$, and the landmark uniqueness value of \textit{overlooked} landmarks $\mathcal{NL}_{G}$ into $\Upsilon_{\mathcal{NL}_{G}}$.
Our uniqueness heuristic is denoted as $h_{\widetilde{UNIQ}}$ and formally defined in Equation~3.

\begin{figure}[th!]
  \centering
  \includegraphics[width=0.95\linewidth]{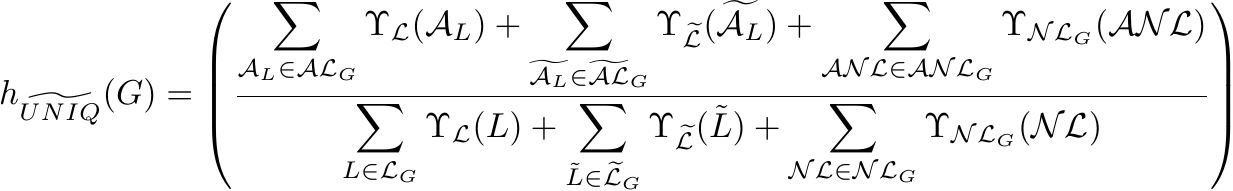}{
  ~(3)}
\end{figure}


\section{Experiments and Evaluation}\label{section:ExperimentsAndEvaluation}

We now describe the experiments carried out to evaluate our goal recognition and discuss the results over our new dataset. 

\subsection{Dataset and Setup}

For experiments, we used openly available goal and plan recognition datasets~\cite{Pereira_Meneguzzi_PRDatasets_2017}, which contain thousands of recognition problems. 
These datasets contain large and non-trivial planning problems (with optimal and sub-optimal plans as observations) for 15 planning domains, including domains and problems from datasets that were developed by Ram{\'{\i}}rez and Geffner~\shortcite{RamirezG_IJCAI2009,RamirezG_AAAI2010}. 
All planning domains in these datasets are encoded using the STRIPS fragment of PDDL. 
Each goal/plan recognition problem in these datasets contains a complete domain definition, an initial state, a set of candidate goals, a correct hidden goal in the set of candidate goals, and an observation sequence. 
An observation sequence contains actions that represent an optimal or sub-optimal plan that achieves a correct hidden goal, and this observation sequence can be full or partial. 
A full observation sequence represents the whole plan that achieves the hidden goal, \idest, 100\% of the actions having been observed. 
A partial observation sequence represents a plan for the hidden goal, varying in 10\%, 30\%, 50\%, or 70\% of its actions having been observed. 
To evaluate our goal recognition approaches in incomplete domain models, we modify the domain models of these datasets by adding annotated possible preconditions and effects. 
Thus, the only modification to the original datasets is the generation of new, incomplete, domain models for each problem, varying the percentage of incompleteness in these domains. 

We vary the percentage of incompleteness of a domain from 20\% to 80\%. 
For example, consider that a complete domain has, for all its actions, a total of 10 preconditions, 10 add effects, and 10 delete effects. 
A derived model with 20\% of incompleteness needs to have 2 possible preconditions (8 known preconditions), 2 possible add effects (8 known add effects), and 2 possible delete effects (8 known delete effects), and so on for other percentages of incompleteness.  
Like~\cite{PlanningIncomplete_NguyenK_2014,Nguyen_AIJ_2017}, we generated the incomplete domain models by following three steps involving randomly selected preconditions/effects: 
(1) move a percentage of known preconditions and effects into lists of possible preconditions and effects; 
(2) add possible preconditions from delete effects that are not preconditions of a corresponding operator; and 
(3) add into possible lists (of preconditions, add effects, or delete effects) predicates whose parameters fit into the operator signatures and are not precondition or effects of the operator. 
These three steps yield three different incomplete domain models from a complete domain model for each percentage of domain incompleteness with different possible lists of preconditions and effects.
%
%
We ran all experiments using a single core of a 12 core Intel(R) Xeon(R) CPU E5-2620 v3 @ 2.40GHz with 16GB of RAM, and a JavaVM with a 2GB memory and a 2-minute time limit. 

\subsection{Evaluation Metrics}

We evaluated our approaches using three metrics: recognition time in seconds (\textit{Time}); \textit{Accuracy} (\textit{Acc} \%)\footnote{This metric is analogous to the \textit{Quality} (Q) metric , used for most planning-based goal recognition approaches~\cite{RamirezG_IJCAI2009,RamirezG_AAAI2010,NASA_GoalRecognition_IJCAI2015,Sohrabi_IJCAI2016}.}, representing the rate at which the the algorithm returns the correct goal $G^*$ among the most likely goals in $\mathcal{G}$; and \textit{Spread in} $\mathcal{G}$ (\textit{S}) as the average number of returned goals.
As each percentage of domain incompleteness has three different incomplete domain models, the percentage columns (20\%, 40\%, 60\%, and 80\%) in Table~\ref{tab:ExperimentalResults} report averages for \textit{Time}, \textit{Acc} \%, and \textit{S} by taking into account the results of the three incomplete domain models. 
The first column of Table~\ref{tab:ExperimentalResults} shows the total number of goal recognition problems for each domain name, and each row expresses averages for the number of candidate goals $|\mathcal{G}|$; the percentage of the plan that was actually observed (\% \textit{Obs}); and the average number of observations per problem $|Obs|$.

We adapt the \textit{Receiver Operating Characteristic} (ROC) curve metric to show the trade-off between true positive and false positive results.
An ROC curve is often used to compare the true positive predictions as well as the false positive predictions of the experimented approaches. 
Here, each prediction result of our goal recognition approaches represents one point in the space, and thus, instead of a curve, our graphs show the spread of our results over ROC space. 
In the ROC space, the diagonal line represents a random guess to recognize a goal from observations. 
This diagonal line divides the ROC space in such a way that points above the diagonal represent good classification results (better than random guess), whereas points below the line represent poor results (worse than random guess). 
The best possible (perfect) prediction for recognizing goals are points in the upper left corner (0,100). 

\subsection{Results}

\begin{table*}[p]
\centering
\setlength\tabcolsep{2.5pt}
\fontsize{5}{6}\selectfont
\begin{tabular}{|c|c|cc|ccc|ccc|ccc|ccc|ccc|ccc|ccc|ccc|}
\hline
\multicolumn{4}{|c|}{{\textit{Incompleteness of} $\mathcal{\widetilde{D}}$ (\%)}} & \multicolumn{6}{c|}{20\%}                                       & \multicolumn{6}{c|}{40\%}                                       & \multicolumn{6}{c|}{60\%}                                       & \multicolumn{6}{c|}{80\%}                                  \\ \hline
\multicolumn{4}{|c|}{}               & \multicolumn{3}{c|}{$h_{\widetilde{GC}}$}       & \multicolumn{3}{c|}{$h_{\widetilde{UNIQ}}$} & \multicolumn{3}{c|}{$h_{\widetilde{GC}}$}       & \multicolumn{3}{c|}{$h_{\widetilde{UNIQ}}$} & \multicolumn{3}{c|}{$h_{\widetilde{GC}}$}       & \multicolumn{3}{c|}{$h_{\widetilde{UNIQ}}$} & \multicolumn{3}{c|}{$h_{\widetilde{GC}}$}  & \multicolumn{3}{c|}{$h_{\widetilde{UNIQ}}$} \\ \hline
\#        & $|\mathcal{G}|$    & \textit{Obs} (\%)    & $|Obs|$    & \textit{Time} & \textit{Acc} \% & \textit{S} & \textit{Time}  & \textit{Acc} \%  & \textit{S} & \textit{Time} & \textit{Acc} \% & \textit{S} & \textit{Time}  & \textit{Acc} \%  & \textit{S} & \textit{Time} & \textit{Acc} \% & \textit{S} & \textit{Time}    & \textit{Acc} \%   & \textit{S}   & \textit{Time} & \textit{Acc} \% & \textit{S} & \textit{Time}    & \textit{Acc} \%   & \textit{S}   \\
\hline
\end{tabular}
\caption{Experimental results of our approaches for recognizing goals in incomplete STRIPS domain models.}
\label{tab:ExperimentalResults}
\end{table*}
\begin{figure*}[p]
\centering
\begin{minipage}[t]{0.22\linewidth}
	\centering
    \includegraphics[width=1\linewidth]{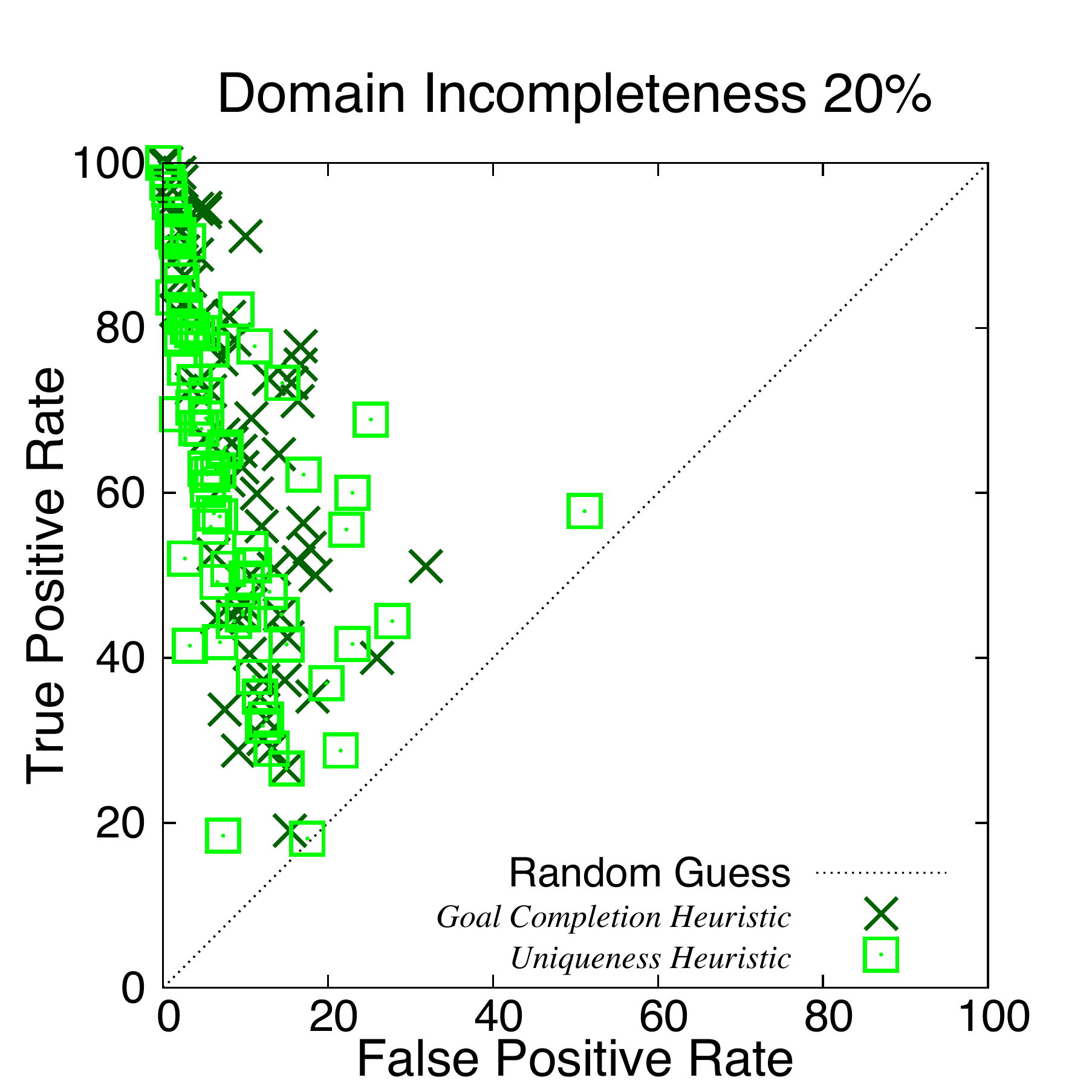}
	\label{fig:ROC_Curve_20}
\end{minipage}
\begin{minipage}[t]{0.22\linewidth}
	\centering
    \includegraphics[width=1\linewidth]{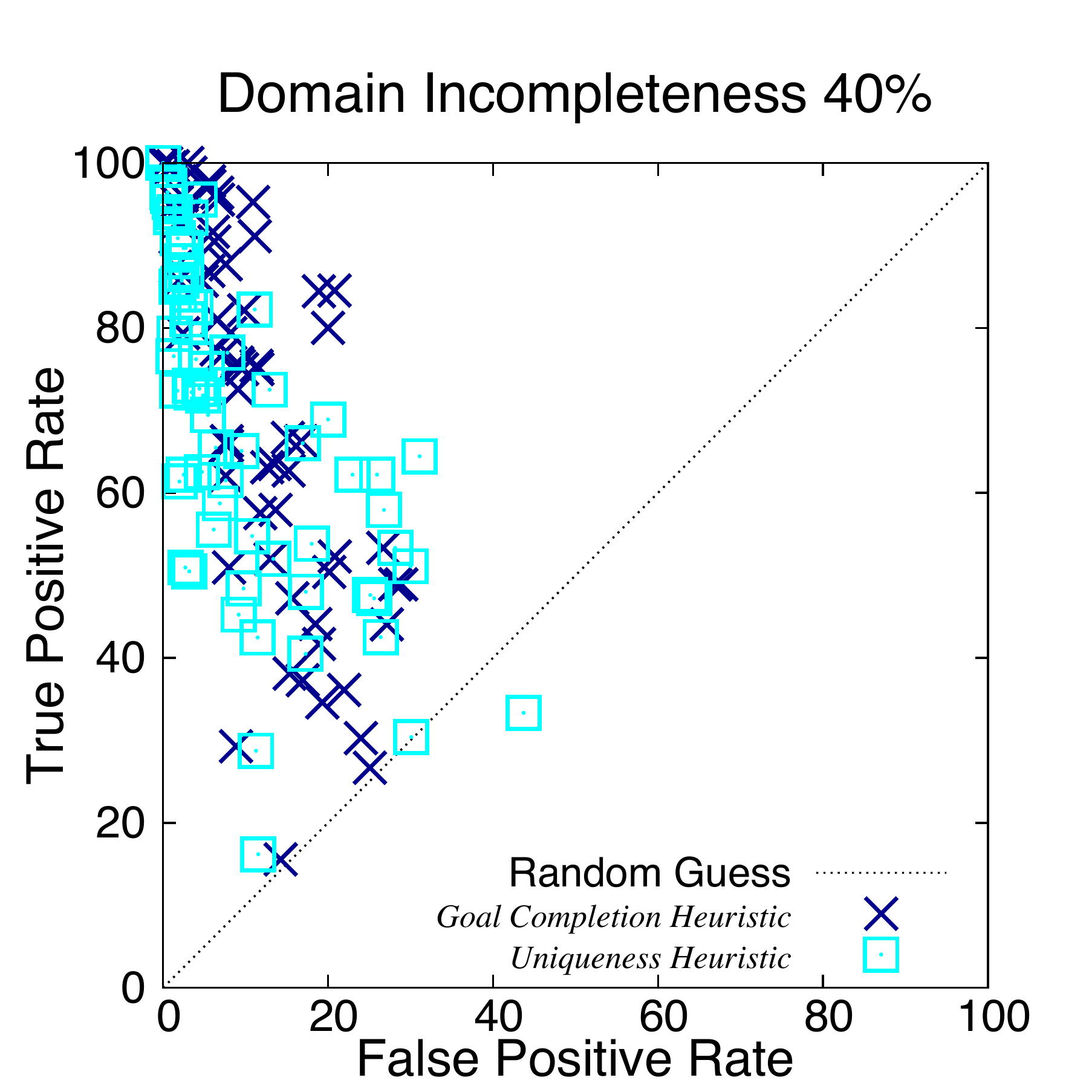}
	\label{fig:ROC_Curve_40}
\end{minipage}
\begin{minipage}[t]{0.22\linewidth}
	\centering
    \includegraphics[width=1\linewidth]{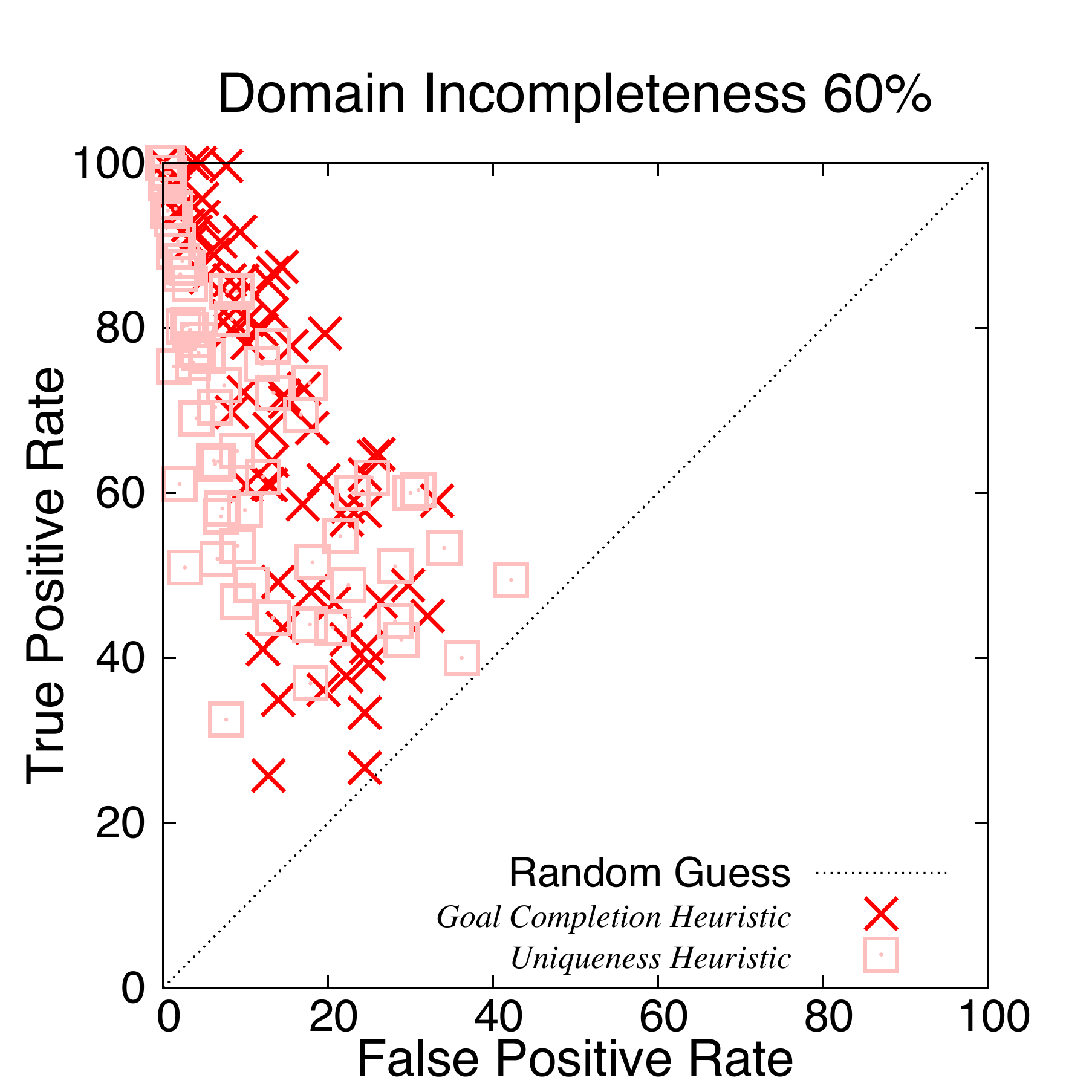}
	\label{fig:ROC_Curve_60}
\end{minipage}
\begin{minipage}[t]{0.22\linewidth}
	\centering
    \includegraphics[width=1\linewidth]{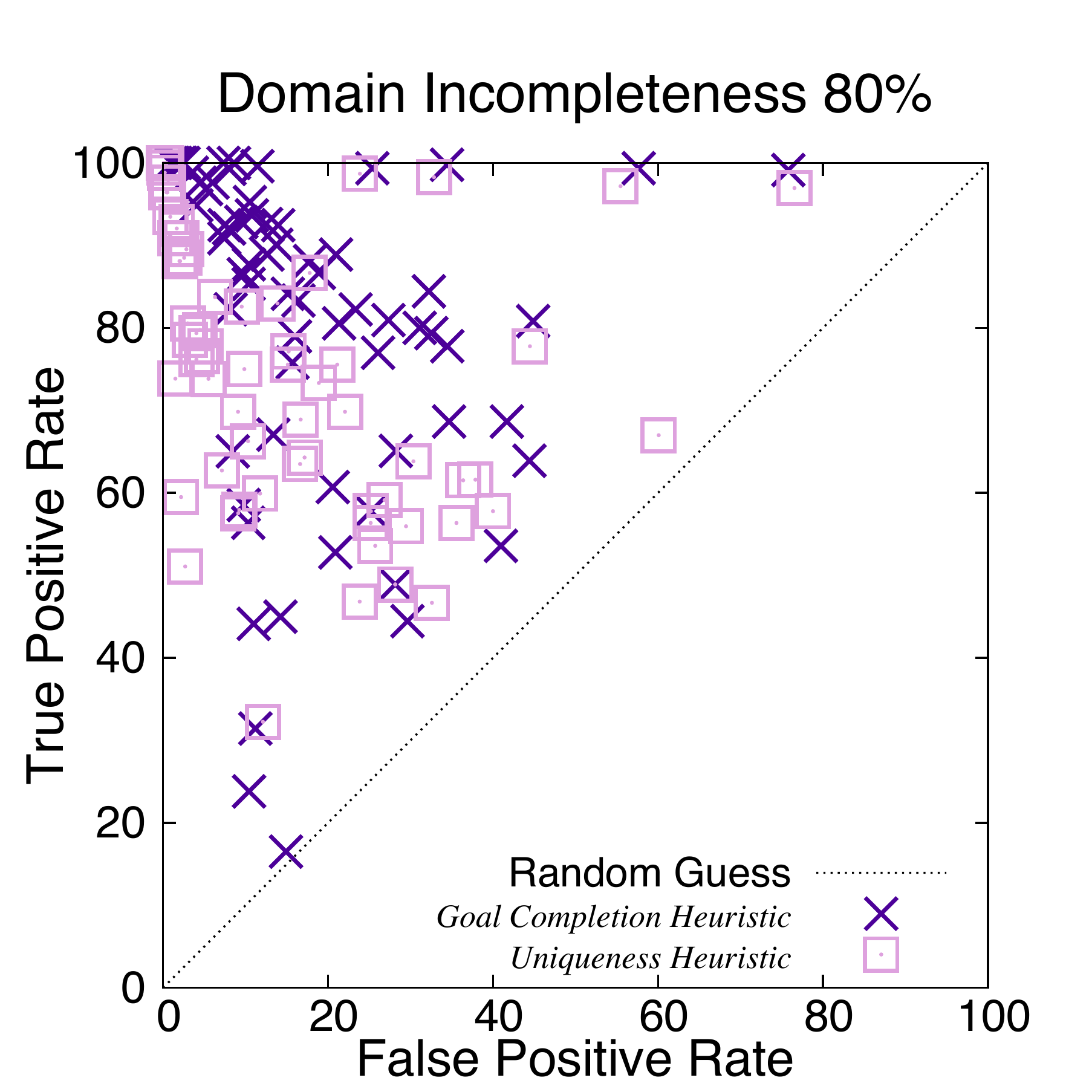}
	\label{fig:ROC_Curve_80}
\end{minipage}
\caption{ROC space for all four percentage of domain incompleteness.}
\label{fig:ROC_Curve_AllDomains}
\end{figure*}

Table~\ref{tab:ExperimentalResults} shows the experimental results of our goal recognition approaches in incomplete domain models. 
Our approaches yield high accuracy at low recognition time for most planning domains apart from \textsc{IPC-Grid} and \textsc{Sokoban}, which took substantial recognition time. 
\textsc{Sokoban} exceeds the time limit of 2 minutes for most goal recognition problems because this dataset contains large problems with a huge number of objects, leading to an even larger number of instantiated predicates and actions. 
For example, as domain incompleteness increases (\idest, the ratio of possible to definite  preconditions and effects), the number of possible actions (moving between cells and pushing boxes) in a grid with 9x9 cells and 5 boxes increases substantially because as there are very few definite preconditions for several possible preconditions. 
The average number of possible complete domain models $|\langle\langle \widetilde{\mathcal{D}} \rangle\rangle|$ is huge for several domains, showing that the task of goal recognition in incomplete domains models is quite difficult and complex.

Figure~\ref{fig:ROC_Curve_AllDomains} shows four ROC space graphs corresponding to recognition performance over the four percentages of domain incompleteness we used in our experiments. 
We aggregate multiple recognition problems for all domains and plot these results in ROC space varying the percentage of domain incompleteness. 
Although the true positive rate is high for most recognition problems at most percentages of domain incompleteness, as the percentage of domain incompleteness increases, the false positive rate also increases, leading to several problems being recognized with a performance close to the random guess line. 
This happens because the number of extracted landmarks decreases significantly as the number of definite preconditions and effects diminishes, and consequently, all candidate goals have few (if any) landmarks. 
For example, in several cases in which domain incompleteness is 60\% and 80\%, the set of landmarks is quite similar, leading our approaches to return more than one candidate goal (increasing the \textit{Spread} in $\mathcal{G}$) as the correct one. 
Thus, there is more uncertainty in the result of the recognition process as incompleteness increases.

\section{Conclusions and Future Work}

We have developed a novel goal recognition approaches that deal with incomplete domain models that represent \textit{possible} preconditions and effects besides traditional models where such information is assumed to be \textit{known}. 
The main contributions of this paper include the formalization of goal recognition in incomplete domains, two heuristic approaches for such goal recognition, novel notions of landmarks for incomplete domains, and a dataset to evaluate the performance of such approaches. 
Our novel notions of landmarks include that of possible landmarks for incomplete domains as well as overlooked landmarks that allow us to compensate fast but non-exhaustive landmark extraction algorithms, the latter of which can also be employed to improve existing goal and plan recognition approaches~\cite{PereiraMeneguzzi_ECAI2016,PAIR17_PereiraOrenMeneguzzi,PereiraNirMeneguzzi_AAAI2017}. 
Experiments over thousands of goal recognition problems in 15 planning domain models show that our approaches are fast and accurate when dealing with incomplete domains at all variations of observability and domain incompleteness.

Although our results are significant, this is just the first work to solve the problem of goal recognition in incomplete domains and a number of refinements can be investigated as future work. 
Since our approaches only explore the set of possible add effects to build an ORPG, we can further improve our approaches by exploring the set of possible preconditions for extracting landmarks. 
Second, we intend to use a propagated RPG to reason about impossible incomplete domain models, much like in~\cite{WeberBryce_ICAPS_2011}, to build a planning heuristic. 
Third, we aim to use a bayesian framework to compute probabilistic estimations of which possible complete domain is most consistent with the observations. 
Fourth, we can use recent work on the refinement of incomplete domain models based on plan traces~\cite{RefiningSTRIPSIncomplete_Zhuo_2013} as part of a complete methodology to infer domains with incomplete information based on plan traces. 
Finally, and most importantly, being able to recognize goals in incomplete domains allows us to employ learning techniques to automated domain modeling~\cite{Jimenez:2012wl} and cope with its possible inaccuracies.

\bibliographystyle{named}
\bibliography{ijcai18-gr-incomplete_domain_models}

\end{document}